\documentclass[10pt,twocolumn,letterpaper]{article}

\usepackage{cvpr}              
\usepackage{multirow}
\usepackage{tabularx}

\definecolor{cvprblue}{rgb}{0.21,0.49,0.74}
\usepackage[pagebackref,breaklinks,colorlinks,allcolors=cvprblue]{hyperref}
\usepackage{caption}
\usepackage{color}

\def\paperID{2722} 
\def\confName{CVPR}
\def\confYear{2025}

\title{TiFRe: Text-guided Video Frame Reduction for \\ Efficient Video Multi-modal Large Language Models}

\author{Xiangtian Zheng\textsuperscript{1}\;\;\;\;\;\;Zishuo Wang\textsuperscript{1}\;\;\;\;\;\;Yuxin Peng\textsuperscript{1*}\\
\textsuperscript{1}Wangxuan Institute of Computer Technology, Peking University\\
{\tt\small 2100013103@stu.pku.edu.cn, wangzishuo@pku.edu.cn, pengyuxin@pku.edu.cn}
}

\begin{document}
\maketitle
\begin{abstract}

With the rapid development of Large Language Models (LLMs), Video Multi-Modal Large Language Models (Video MLLMs) have achieved remarkable performance in video-language tasks such as video understanding and question answering. However, Video MLLMs face high computational costs, particularly in processing numerous video frames as input, which leads to significant attention computation overhead. A straightforward approach to reduce computational costs is to decrease the number of input video frames. However, simply selecting key frames at a fixed frame rate (FPS) often overlooks valuable information in non-key frames, resulting in notable performance degradation. To address this, we propose Text-guided Video Frame Reduction (TiFRe), a framework that reduces input frames while preserving essential video information. TiFRe uses a Text-guided Frame Sampling (TFS) strategy to select key frames based on user input, which is processed by an LLM to generate a CLIP-style prompt. Pre-trained CLIP encoders calculate the semantic similarity between the prompt and each frame, selecting the most relevant frames as key frames. To preserve video semantics, TiFRe employs a Frame Matching and Merging (FMM) mechanism, which integrates non-key frame information into the selected key frames, minimizing information loss. Experiments show that TiFRe effectively reduces computational costs while improving performance on video-language tasks.

\end{abstract}
    
\section{Introduction}
\label{sec:intro}

\begin{figure}[h]
  \centering
  \includegraphics[width=\linewidth]{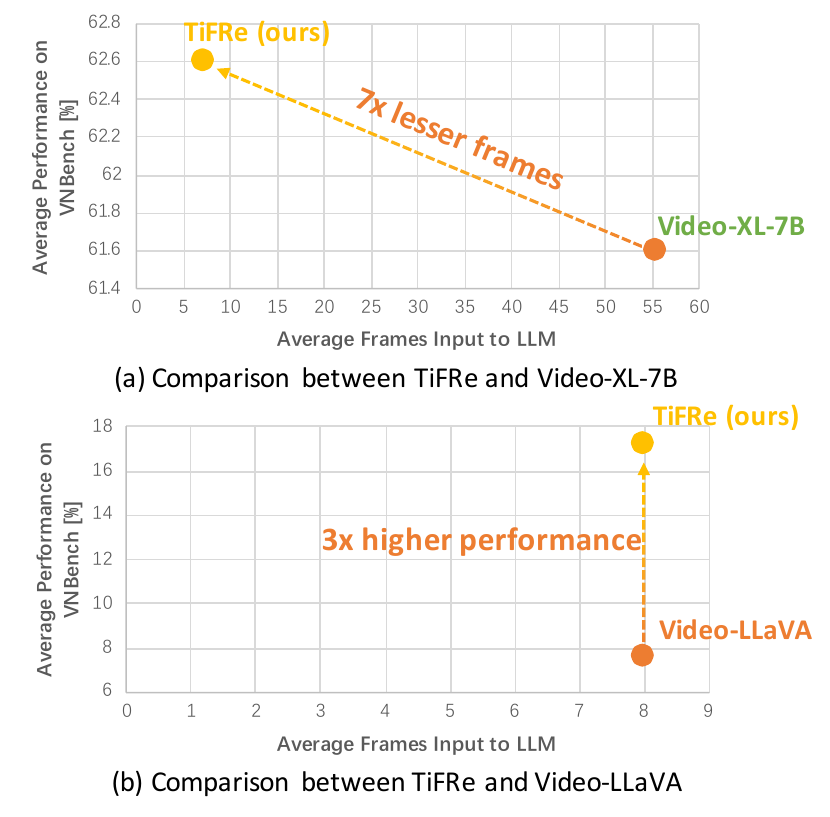}
  \caption{The performance comparison between our proposed method TiFRe and State-of-the-Art Video MLLM Video-XL-7B \cite{shu2024video} and Video-LLaVA \cite{lin2023video} on VNBench. Compared with Video-XL, TiFRe achieves a significant reduction in input frames while obtaining a higher performance. Compared with Video-LLaVA, TiFRe achieves a substantial performance improvement with the same number of input frames.
}
  \label{fig:intro-1}
\end{figure}

\begin{figure*}[t]
    \centering
    \includegraphics[width=1\textwidth]{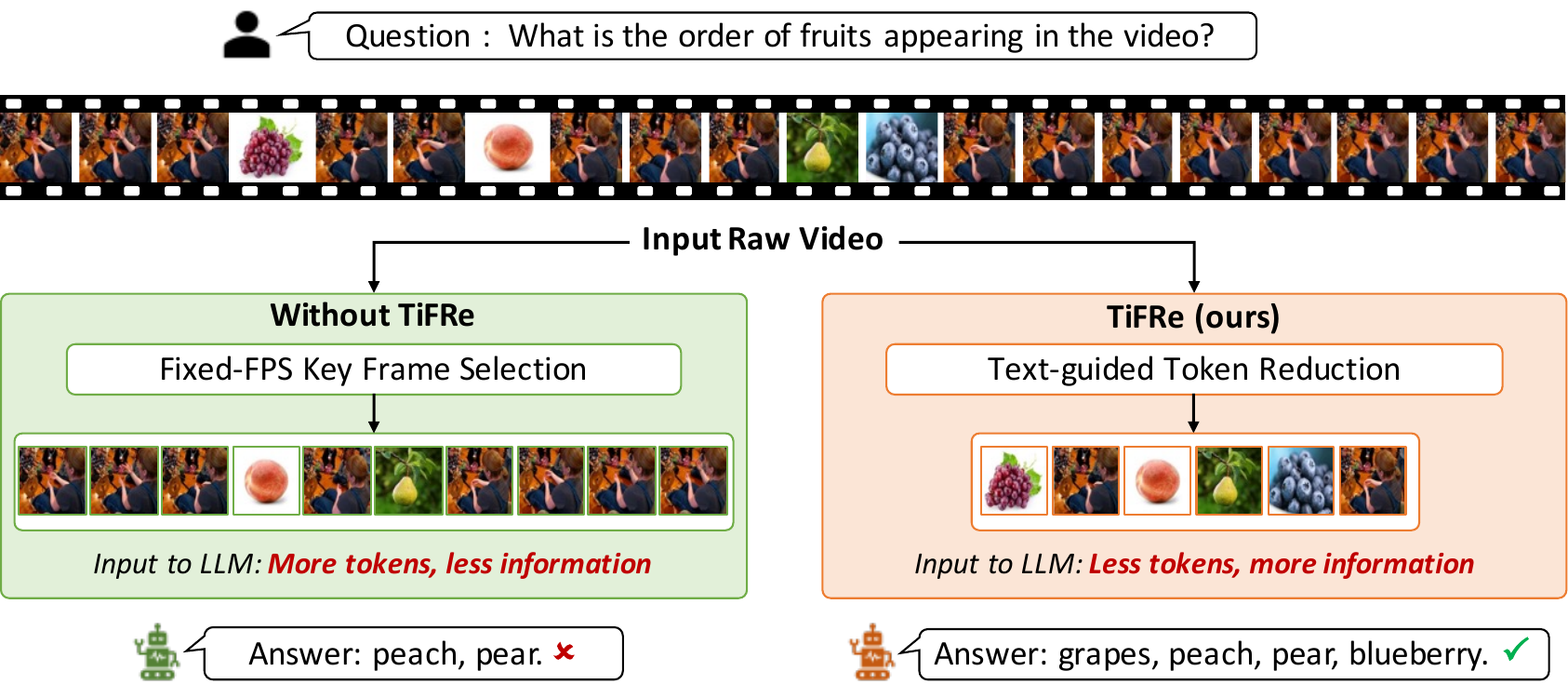}
    \caption{This illustration compares two different frame selection methods used in Video MLLMs. On the left, the Fixed-FPS Key Frame Selection method samples a fixed number of frames evenly across the video, which may result in losing semantics and redundant tokens. On the right, the Text-guided Frame Reduction (TiFRe) selects frames with significant semantic information, especially those relevant to the text input. The results indicate that TiFRe's frame selection strategy better maintains the video information and eliminates redundant frames, leading to a more accurate and effective response.}
    \label{fig:intro-2}
    \vspace{-0.3cm}
\end{figure*}

With significant advances in Large Language Models (LLMs) \cite{touvron2023llama, mann2020language} for natural language processing, Multimodal Large Language Models (MLLMs) \cite{liu2024visual} have demonstrated remarkable performance in computer vision and multimodal tasks.
MLLMs align the visual embeddings obtained from a pre-trained visual encoder \cite{radford2021learning} with the text embeddings, and then jointly input them into the transformers of LLMs, achieving perception and reasoning of modalities such as images and videos. Among the MLLMs, Video MLLMs \cite{lin2023video, li2023videochat, wang2024internvideo2, zhang2024video} show competitive performance on various tasks, including action recognition, video question answering, and video summarization \cite{tang2023video}.

However, since videos generally have a higher frame rate, Video MLLMs need to input the image features of these frames together into the LLM transformers, leading to longer token sequences. This makes the inference computational cost of video MLLM significantly higher than that of LLMs and image MLLMs \cite{park2024too, li2024llms}. The challenge of high inference costs becomes more severe when dealing with long videos, but not all frames in a video are irreplaceable. Due to the continuity of videos, information redundancy exists between adjacent frames. Some Video MLLMs reduce the number of frames input to LLM by extracting key frames, thereby reducing the inference costs \cite{tan2024koala, liang2024keyvideollm}.

When selecting key frames from the input video, the key challenge is \textbf{how to select as few key frames as possible without losing information}. Based on the observation that, key frames should be selected according to different prompts input into the Video MLLM. As shown in Figure \ref{fig:intro-2}, when the input question is "What is the order of fruits appearing in the video?", the key frames should be the frames containing fruits. Therefore, we divide the challenge of key frame selection into two sub-problems: (1) How to select as few key frames as possible based on different prompts? (2) After selecting key frames, how to recover the information of other non-key frames as much as possible to avoid information loss?

To address the two sub-problems, we propose a new framework for efficient Video MLLMs named \textbf{\textit{TiFRe}}. TiFRe selects key frames from raw videos based on the user’s prompt and then fuses information from non-key frames into the selected frames to minimize information loss. The framework comprises two main steps: (1) \textbf{Text-guided Frame Sampling (TFS)} and (2) \textbf{Frame Matching and Merging (FMM)}.TFS aims to identify target frames based on the user’s input prompt, selecting frames that contain the relevant objects as key frames. The process begins by combining the user’s prompt with an additional manually defined prompt, which are input into an LLM. The LLM extracts the names of the target objects and generates a CLIP-style prompt in the format of “a photo of a [target object].” Pre-trained CLIP image and text encoders are then used to calculate the semantic similarity between the CLIP-style prompt and each video frame. Frames with the highest similarity scores are selected as key frames.FMM fuses information from non-key frames into the selected key frames to preserve the overall video semantics. This is achieved by matching each unselected frame to its most similar key frame.For each key frame, FMM computes the weighted average of the frames assigned to it, along with the key frame itself, to generate the final key frame input. 

Figure \ref{fig:intro-1} compares the performance of TiFRe with two existing Video MLLMs: Video-XL-7B \cite{shu2024video} and Video-LLaVA \cite{lin2023video}. Both models sample key frames at a fixed frame rate from the raw video. In the case of Video-XL, the higher frame rate results in redundant frames. TiFRe addresses this by reducing the number of frames input to the MLLM from 55.2 to 8.6, significantly lowering inference costs while achieving even higher accuracy. For Video-LLaVA, only 8 key frames are selected per video, leading to severe information loss. In contrast, TiFRe retains the same number of key frames while improving accuracy from 7.6 to 19.2.

The contributions of this paper can be summarized as follows: 

(1) \textbf{Proposal of a Novel Framework}: We propose TiFRe, a text-guided frame reduction framework specifically designed to enhance the efficiency of Video Multimodal Large Language Models (Video MLLMs).

(2) \textbf{Prompt-Based Key Frame Selection}: TiFRe dynamically selects key frames based on user prompts and integrates unselected frame information to preserve video semantics with minimal information loss.

(3) \textbf{Improved Efficiency and Accuracy}:TiFRe effectively reduces the number of input frames required for processing, leading to a substantial reduction in computational costs while achieving notable improvements in accuracy.
\section{Related Work}
\label{sec:related}
\subsection{Video Multi-modal LLMs}

Video Multi-modal Large Language Models (MLLMs) represent an innovative and rapidly evolving class of models designed to process and interpret video embeddings extracted from pre-trained visual encoders, alongside user-provided textual prompts. These combined multi-modal inputs are fed through large language model (LLM) transformers, which decode them into meaningful text outputs, thereby bridging the gap between visual and linguistic modalities. This approach enables MLLMs to achieve sophisticated levels of video content understanding, supporting effective video analysis, comprehension, and interaction. By integrating visual and language processing, MLLMs address a variety of complex video-related tasks, making them powerful tools for both analysis and creative applications. Moreover, their ability to understand video context and provide natural language responses makes them ideal for interactive AI solutions, enhancing user experience in domains like education, entertainment, and customer service. One notable model is VideoLLM \cite{chen2023videollm}, which builds a versatile framework capable of handling a wide range of video understanding tasks by converting multi-modal inputs into a unified token sequence, thereby enhancing adaptability and robustness. Similarly, Video-LLaMA \cite{zhang2023video} employs a video Q-former to transform video embeddings into token sequences, while incorporating auditory information via ImageBind \cite{girdhar2023imagebind} to further enhance its multi-modal understanding capabilities.

Other significant advancements include Video-ChatGPT \cite{maaz2023video}, which combines a video-specific encoder with an LLM to facilitate interactive, video-grounded conversations, enhancing user engagement and accessibility. MovieChat \cite{song2024moviechat} introduces a memory mechanism to tackle the challenge of processing long videos, effectively retaining temporal dependencies and maintaining narrative context over extended sequences. Additionally, Video-LLaVA \cite{lin2023video} aligns visual and linguistic feature spaces to jointly learn from both static and dynamic content, thus benefiting from cross-modal reinforcement. Extending beyond understanding, GPT4Video \cite{wang2024gpt4video} integrates visual encoders, LLMs, and generative models to excel in both video comprehension and generation tasks, showcasing the vast potential of unified multi-modal frameworks for video-based AI applications. These advancements illustrate how MLLMs are evolving to provide richer and more interactive ways to engage with video content, setting the stage for further breakthroughs in AI-driven video solutions.

While these video MLLMs demonstrate exceptional performance across a variety of tasks, their computational demands are significantly higher than image-based LLMs due to the large number of video frames involved in processing. This increased complexity poses challenges for scalability, especially in resource-intensive applications. Therefore, minimizing the number of input frames without compromising the richness of video information is critical to making these models more efficient and practical for real-world deployments.

\subsection{Frame Reduction for Video MLLMs}

Frame reduction is a crucial strategy for optimizing the computational efficiency of video MLLMs, as it addresses the problem of excessive token sequences by reducing the number of input frames while retaining essential information. A commonly used approach is to sample key frames from the raw video at a fixed frame rate \cite{zhang2023video, wang2024internvideo2, shu2024video}. This method is straightforward and computationally efficient, but it suffers from a significant drawback: the fixed sampling rate can result in the omission of critical frames, particularly in videos with uneven temporal dynamics, leading to a degradation in model performance on tasks that require nuanced comprehension of video content.

To overcome the limitations of fixed sampling, more advanced frame reduction techniques have been proposed. Video Token Sparsification (VTS) \cite{ma2024video} introduces a lightweight convolutional neural network (CNN) to identify and retain key frames while pruning redundant information in consecutive frames. By intelligently analyzing frame content, VTS reduces the total number of video tokens without losing critical semantic or visual information, offering an effective balance between computational efficiency and performance. Building on this, Video Token Merging (VTM) \cite{lee2024video} enhances frame reduction by incorporating saliency-based token merging. Unlike similarity-based approaches, VTM dynamically merges tokens based on their saliency, assigning greater importance to tokens representing visually or semantically significant objects while de-emphasizing less relevant elements such as static backgrounds. This approach ensures that the reduced token sequence retains high informational value, even with fewer tokens.

Additionally, Video-ChatGPT \cite{maaz2023video} adopts a machine learning-based key frame selection tool called Katna to identify and extract the most informative frames from raw video data. This technique further refines the input token sequence, allowing the model to process only the most relevant portions of the video, thereby streamlining computational requirements without sacrificing task performance. Together, these frame reduction techniques represent a robust suite of tools for managing the complexity of video MLLMs, enabling these models to process video content more efficiently while maintaining or even enhancing their ability to perform on complex video-related tasks.

The adoption of frame reduction methods not only reduces computational costs but also enhances the scalability of video MLLMs for real-world applications. By pruning redundancy and retaining essential video content, these techniques make it feasible to deploy MLLMs for large-scale video understanding tasks, such as long-form video summarization, real-time video-based interaction, and resource-constrained settings where computational efficiency is paramount. This line of research continues to be a critical enabler for advancing the practical applicability of video MLLMs.


\begin{figure*}[t]
    \centering
    \includegraphics[width=1\textwidth]{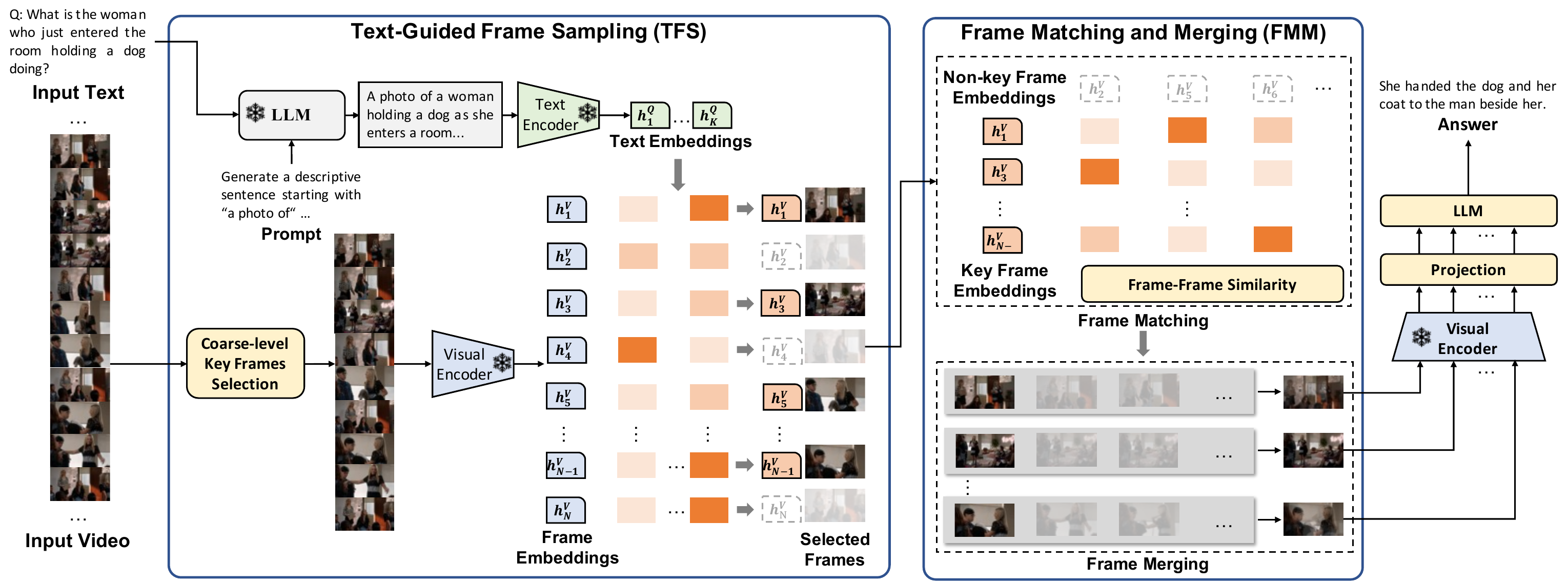}
    \caption{The framework of our proposed Text-guided Video Frame Reduction (TiFRe).}
    \label{fig:framework}
    \vspace{-0.3cm}
\end{figure*}

\section{Method}

\subsection{Task Definition and Overview}

This paper primarily focuses on the general video question answering task. Given a video $\mathcal{V}$ and a textual question $\mathcal{Q}$, the main objective of TiFRe is to strategically reduce the number of frames input to the Multimodal Large Language Model (MLLM). By selecting a representative subset of frames, TiFRe aims to decrease the computational cost associated with processing the entire video, thereby enhancing efficiency without compromising performance. The objective of TiFRe is formally expressed as:
\begin{equation}
    \{\hat{I}_1,\ldots, \hat{I}_M\} = TiFRe(\mathcal{V}).
\end{equation}

where ${\hat{I}_1,\ldots, \hat{I}_M}$ represents the $M$ selected key frames from the video $\mathcal{V}$. These frames act as a distilled representation of the video, allowing the MLLM to focus on the most critical content while answering the question $\mathcal{Q}$ efficiently and effectively.

In this way, TiFRe provides a scalable and adaptive solution to the computational challenges of VideoQA, making it well-suited for processing long or complex videos without sacrificing semantic fidelity.

\subsection{Text-Guided Frame Sampling}

Motivated by the observation that, given the same video, different key frames should be selected based on different user prompts, we aim to extract the target objects specified in the user prompts and select key frames accordingly. This approach emphasizes the alignment of user intent with frame selection, ensuring that the extracted frames are semantically meaningful and relevant to the provided descriptions.

One straightforward approach to achieve this is to leverage the powerful pre-trained CLIP model, which is designed to compute image-text similarity effectively. CLIP bridges the gap between textual and visual domains by mapping both to a shared embedding space. However, user inputs are often presented in various forms, such as questions or free-form natural language descriptions, which may not directly conform to the structured text prompt format required by CLIP.  Therefore, we first input the question $\mathcal{Q}$ to an LLM with the prompt $\mathcal{P}$:
\begin{quote}
    \textit{Generate a descriptive sentence starting with "a photo of" based solely on the key nouns from the following question and its options. Exclude words like "video," "order," and similar terms. The sentence should be concise and describe a single frame image for CLIP's image search.
    \\ Question:}
\end{quote}
The user question $\mathcal{Q}$ is input together with the prompt $\mathcal{P}$ into the LLM. The LLM will output a set of text prompts $\{\mathcal{T}_1,...,\mathcal{T}_K\}$ following the CLIP style, which start with "a photo of a". $K$ is the number of target objects extracted from the user question. The CLIP text encoder is denoted as $F_\mathcal{T}$, so the text embedding of the user question $h^\mathcal{Q}$ can be formulated as:
\begin{equation}
    h^\mathcal{Q}_i = F_\mathcal{T}(\mathcal{T}_i) = F_\mathcal{T}(LLM([\mathcal{P}, \mathcal{Q}])_i).
\end{equation}

For the input video $\mathcal{V}$, we first adopt a coarse-level key frame selection, extracting 1 frame every 1 second from the raw video. Suppose there are $N$ frames extracted from the raw video $\mathcal{V}$, denoted as $\{I_1,\ldots, I_N\}$. We then use the CLIP image encoder $F_\mathcal{V}$ to extract the visual features of the $N$ frames. The visual feature of the $j$th frames $h^\mathcal{V}_j$ if denoted as:
\begin{equation}
    h^\mathcal{V}_j = F_\mathcal{V}(I_j).
\end{equation}

This step converts each video frame into a high-dimensional embedding that resides in the same shared space as the text embeddings, enabling direct comparison between textual and visual representations.After obtaining the text features of the user question and the visual features of $N$ video frames, we compute the cosine similarity between the $i$th text feature and the $j$th visual feature:
\begin{equation}
    SIM_{ij} = \frac{h_i^{\mathcal{Q}} \cdot h_j^{\mathcal{V}}}{\|h_i^{\mathcal{Q}}\| \|h_j^{\mathcal{V}}\|}.
\end{equation}
The saliency score for the $j$ frame $S_j$ is calculated as the maximum of similarities:
\begin{equation}
    S_j = \max _{i=1,\ldots,K}\ SIM_{ij}.
\end{equation}
We select the top $M$ frames with the highest saliency scores as $M$ key frames, denoted as $\{I'_1,\ldots,I'_M\}$. These selected frames represent the most semantically relevant moments in the video as determined by the user prompt. This adaptive and context-aware selection process ensures that the resulting key frames align closely with the user's intent.

\subsection{Frame Matching and Merging}

After extracting $M$ key frames from the raw video, it is crucial to ensure that the selected frames can represent the overall semantics of the video as comprehensively as possible. While the key frames capture the most salient moments based on the user prompt, the remaining frames also contain valuable contextual information that contributes to a richer understanding of the video. Therefore, our approach involves retaining the $M$ key frames while incorporating information from the non-key frames to refine and enhance the representation of these key frames.

We denote the set of indexes corresponding to the key frames as $A_k$, and the indexes corresponding to the remaining non-key frames as $A_n$. The key frames are represented as ${I_k, k \in A_k}$, and the non-key frames are represented as ${I_k, k \in A_n}$. These two sets together constitute all the frames extracted from the video:
\begin{equation}
    \{I_k, k \in A_k\} \bigcup \{I_k, k \in A_n\} = \{I_1,\ldots, I_N\}.
\end{equation}
For each non-key frame $I_k, k\in A_n$, we conduct frame matching by finding the most similar key frame $I_{m(k)}$. $m$ means the matching operation, which should satisfy:
\begin{equation}
    m(k) = \arg\min _{i \in A_k} \frac{h_i^{\mathcal{V}} \cdot h_k^{\mathcal{V}}}{\|h_i^{\mathcal{V}}\| \|h_k^{\mathcal{V}}\|}.
\end{equation}

This ensures that each non-key frame $I_k$ is assigned to the key frame $I_{m(k)}$ that is most semantically similar based on the shared embedding space between text and visual features.

After frame matching, we conduct frame merging by calculating the average frame of each key frame with non-key frames assigned with it. To be specific, we keep a set $S^F_k$ for each key frame $I_k, k\in A_k$. From each pair $<I_n, I_k>, n\in A_n, k \in A_k$ obtained from the frame matching, we assign the non-key frame to the key frame's set $S^F_k$. Then we calculate the weighted average frame of each set:
\begin{equation}
    \bar{I_k} = \frac{1}{|S^F_k|} \sum _{I_n \in S^F_k} w_n I_n,
\end{equation}
where $w_n$ is the weight for $I_n$, indicating the saliency of the non-key frame. We take the similarity between the non-key frame $I_n$ and the corresponding key frame $I_k$ as the weight:
\begin{equation}
    w_n = \frac{h_n^{\mathcal{V}} \cdot h_k^{\mathcal{V}}}{\|h_n^{\mathcal{V}}\| \|h_k^{\mathcal{V}}\|}.
\end{equation}
The final $M$ key frames can be denoted as:
\begin{equation}
    \{\bar{I_k}, k\in A_k\} = \{\hat{I}_1,\ldots,\hat{I}_M\}.
\end{equation}
These refined key frames ${\hat{I}_1, ..., \hat{I}_M}$ are more robust and semantically representative, as they incorporate both the salient moments identified during the initial selection and the contextual information from other frames. This approach ensures that the final key frames provide a comprehensive summary of the video while aligning closely with the user-defined prompt.

\begin{table*}[t]\footnotesize
\renewcommand{\arraystretch}{1.2}
\centering
\addtolength\tabcolsep{-1.8pt} 
\begin{tabularx}{\linewidth}{>{\centering\arraybackslash}p{4cm} | >{\centering\arraybackslash}p{3.0cm} | >{\centering\arraybackslash}X >{\centering\arraybackslash}X >{\centering\arraybackslash}X >{\centering\arraybackslash}X | >{\centering\arraybackslash}X}
\toprule
\multicolumn{1}{c|}{\multirow{2}{*}{Model}} & \multicolumn{1}{c|}{\multirow{2}{*}{LLM}} & \multicolumn{5}{c}{VNBench} \\
\multicolumn{1}{c|}{} & \multicolumn{1}{c|}{} & Retrieval $\uparrow$ & Ordering $\uparrow$ & Counting $\uparrow$ & Average $\uparrow$ & Frames $\downarrow$ \\ \midrule
LLaMA-VID~\cite{li2025llama} & -- & 25.1 & 0.2 & 7.1 & 10.8 & 55.2 \\
VideoChat2~\cite{li2024mvbench} & -- & 32.7 & 0.4 & 15.5 & 20.1 & 16 \\
LLaVA-NeXT-Video~\cite{liu2024llavanext} & -- & 44.2 & 0.4 & 15.5 & 20.1 & 55.2 \\
ST-LLM~\cite{stllm} & -- & 51.3 & 0.0 & 16.7 & 22.7 & 32 \\
\midrule
Video-XL~\cite{shu2024video} & -- & 80.0 & 80.6 & 24.0 & 61.6 & 55.2 \\ 

\addlinespace[1pt]
\textbf{Video-XL+TiFRe (ours)} & Qwen2.5-7B-Instruct &
82.0 & 79.1 & 26.7 & 62.6 & \textbf{8.6} \\
$\Delta$ {$Acc.$ \textcolor{gray}{(vs.~Video-XL)}} &  &
\textcolor{green}{\textbf{+2.0}} &
\textcolor{green}{\textbf{-1.5}} &
\textcolor{green}{\textbf{+2.7}} &
\textcolor{green}{\textbf{+1.0}} &
\textcolor{green}{\textbf{-84.4\%}} \\

\addlinespace[2pt]
\cmidrule(lr){1-7}

\addlinespace[1pt]
\textbf{Video-XL+TiFRe (ours)} & OpenPangu-7B &
\textbf{82.7} & \textbf{81.6} & \textbf{26.9} & \textbf{63.7} & \textbf{8.6} \\
$\Delta$ {$Acc.$ \textcolor{gray}{(vs.~Video-XL)}} &  &
\textcolor{green}{\textbf{+2.7}} &
\textcolor{green}{\textbf{+1.0}} &
\textcolor{green}{\textbf{+2.9}} &
\textcolor{green}{\textbf{+2.1}} &
\textcolor{green}{\textbf{-84.4\%}} \\
\bottomrule
\end{tabularx}
\vspace{2mm}
\caption{Experimental results on VNBench.}
\label{tab:vnbench}
\end{table*}

\begin{table}[ht]
\centering
\footnotesize
\renewcommand{\arraystretch}{1.3}
\setlength{\tabcolsep}{8pt}
\begin{tabular}{l|c|c|c}
\toprule
\multicolumn{1}{c|}{\multirow{2}{*}{Model}} & \multicolumn{3}{c}{MLVU Dev Set} \\ \cmidrule(lr){2-4}
& Size & m\_avg & Frames \\ \midrule
Video-LLAMA-2~\cite{zhang2023video}        & 13B & 35.5 & 16   \\
MA-LMM  ~\cite{malmm2024}             & 7B  & 36.4 & 1000 \\
VTimeLLM  ~\cite{vtimellm}           & 7B  & 41.9 & 100  \\
Qwen-VL-Max  ~\cite{bai2023qwen}        & -   & 42.2 & 16   \\
MiniGPT4-Video  ~\cite{ataallah2024minigpt4}     & 7B  & 44.5 & 90   \\
VideoChat2-Vicuna ~\cite{videochat2}   & 7B  & 44.5 & 16   \\
ShareGPT4Video ~\cite{chen2024sharegpt4video}      & 8B  & 46.4 & 16   \\
VideoLLAMA2-Chat   ~\cite{damonlpsg2024videollama2}   & 7B  & 48.5 & 16   \\
\midrule
Video-LLAVA  ~\cite{lin2023video}        & 7B  & 47.3 & 8    \\
\textbf{Video-LLAVA+TiFRe(ours)}       & 7B  & \textbf{48.9} & \textbf{8}    \\
\bottomrule
\end{tabular}
\caption{Experimental results on MLVU Dev Set.}
\label{tab:mlvu_dev_set}
\end{table}

\begin{table*}[ht]
\centering
\footnotesize
\renewcommand{\arraystretch}{1.3}
\setlength{\tabcolsep}{10pt} 
\begin{tabular*}{\textwidth}{p{0.3\textwidth}|@{\extracolsep{\fill}}cccc}
\toprule
\multicolumn{1}{c|}{\multirow{2}{*}{Model}}  &  \multicolumn{4}{c}{VNBench} \\
\multicolumn{1}{c|}{}  & Retrieval $\uparrow$    & Ordering $\uparrow$ & Counting $\uparrow$ & Average $\uparrow$    \\ \midrule
Video-XL~\cite{shu2024video}  & 80.0 & \textbf{80.6}   & 24.0 & 61.6  \\
Video-XL+TiFRe, \( k = 4 \)  & 82.0 & 76.7 & 23.6 & 60.7 \\
Video-XL+TiFRe, \( k = 5 \)  & 82.4 & 76.4 & 24.0 & 61.0 \\
Video-XL+TiFRe, \( k = 6 \)  & 82.0 & 76.4 & 26.2 & 61.6 \\
Video-XL+TiFRe, \( k = 7 \)  & \textbf{82.9} & 76.2 & 26.7 & 61.9 \\
Video-XL+TiFRe, \( k = 8 \)  & 82.2 & 76.9 & 26.7 & 61.9 \\
Video-XL+TiFRe, \( k = 10 \) & 82.0 & 79.1   & \textbf{26.7} & \textbf{62.6} \\
Video-XL+TiFRe, \( k = 11 \) & 82.2 & 78.9 & 26.2 & 62.4 \\
Video-XL+TiFRe, \( k = 13 \) & 81.6 & 78.2 & 25.6 & 61.7 \\
Video-XL+TiFRe, \( k = 15 \) & 82.0 & 77.3 & 25.6 & 61.6 \\
\bottomrule
\end{tabular*}
\caption{Effect of Varying \( k \) (Maximum Frame Count) on VNBench.}
\label{tab:ablation_with_k}
\end{table*}

\begin{table*}[t]\footnotesize  
\renewcommand{\arraystretch}{1.3}
\centering  
\addtolength\tabcolsep{-1.8pt} 
\begin{tabularx}{\linewidth}{
>{\centering\arraybackslash}p{4cm} |
>{\centering\arraybackslash}X
>{\centering\arraybackslash}X
>{\centering\arraybackslash}X
>{\centering\arraybackslash}X |
>{\centering\arraybackslash}X
}
\toprule
\multicolumn{1}{c|}{\multirow{2}{*}{Model}}  &  \multicolumn{5}{c}{VNBench} \\
\multicolumn{1}{c|}{}  & Retrieval $\uparrow$    & Ordering $\uparrow$ & Counting $\uparrow$ & Average $\uparrow$ & Frames $\downarrow$   \\ \midrule
Baseline (Video-XL)          & 80.0 & \textbf{80.6}   & 24.0 & 61.6 & 55.2\\
Without TFS                   & 44.4 & 44.0  & 18.2 & 35.6  &\textbf{10.0}\\
Without FMM                   & 81.6 & 78.4 & 25.8 & 61.9 &\textbf{8.6} \\
Full Model (Video-XL+TiFRe) & \textbf{82.0} & 79.1 & \textbf{26.7} & \textbf{62.5} &\textbf{8.6} \\
\bottomrule
\end{tabularx}
\caption{Ablation Study Results on VNBench.}
\label{tab:ablation_without_k}
\end{table*}

\section{Experiments}
\subsection{Benchmarks}
To comprehensively evaluate the effectiveness and flexibility of the proposed framework, TiFRe, we conducted extensive experiments on widely recognized benchmarks designed to assess long video understanding. Specifically, we tested TiFRe on VNBench\cite{vnbench} and MLVU\cite{zhou2024mlvu}. These benchmarks encompass a diverse set of tasks, including retrieval, sequence ordering and counting, providing a holistic evaluation of the framework’s ability to process complex and temporally extended video content.

\subsection{Main results}
\subsubsection{Video Benchmark Evaluation}


\textbf{VNBench} is a synthetic benchmark designed to evaluate long-context reasoning. It includes tasks such as retrieval, sequence ordering, and object/event counting, each testing the model's capacity for reasoning over extended video sequences. Table~\ref{tab:vnbench} demonstrates that TiFRe not only significantly reduces the number of input frames but also improves accuracy across tasks. When integrated with Video-XL, TiFRe reduces the input frame count from 55.2 to 8.6 while achieving a higher average score. Notably, this improvement is consistent across different LLMs: replacing Qwen2.5-7B-Instruct\cite{qwen2025qwen25technicalreport} with OpenPangu-7B\cite{chen2025panguembeddedefficientdualsystem} further boosts performance, achieving the best overall average score of 63.7 while keeping the same compact input of 8.6 frames. OpenPangu-7B also provides uniform gains across subtasks (82.7 retrieval, 81.6 ordering, and 26.9 counting), indicating that TiFRe’s effectiveness is not tied to a specific LLM but generalizes well to alternative language models. These results validate the framework’s capability to balance computational efficiency and task performance.

\textbf{MLVU} is a benchmark specifically designed to evaluate video understanding. As shown in Table~\ref{tab:mlvu_dev_set}, models integrated with TiFRe demonstrate substantial performance improvements in Multiple-Choice tasks while maintaining the same number of input video frames, thereby avoiding any additional computational overhead. These results highlight the efficiency and precision of TiFRe, particularly in handling long-term temporal contexts and complex reasoning tasks.

\subsubsection{Implementation Details}
The proposed framework is built on pre-trained Video-LLaVA and Video-XL models without requiring additional training data. The Text-guided Frame Sampling (TFS) module aims to identify semantically relevant keyframes from video sequences based on a user-provided query. For this purpose, we leverage the OpenPangu-7B\cite{chen2025panguembeddedefficientdualsystem} and Qwen-2.5-7B\cite{qwen2025qwen25technicalreport} models to generate semantic representations of the query and utilize the CLIP model (ViT-B/32) as the visual encoder to extract frame features. CLIP's exceptional performance in aligning visual and textual data enables precise computation of the semantic relevance of each frame to the query. The frames are then ranked based on their relevance scores, and the top-ranking frames are retained as keyframes. The number of retained keyframes is controlled by a hyperparameter \( k \), which defines the maximum allowable number of keyframes. To further optimize frame representation, the Frame Matching and Merging (FMM) module uses CLIP to extract feature vectors from both keyframes and non-keyframes and calculates pairwise similarity scores. All experiments were conducted on a high-performance computing server equipped with 10 NVIDIA A40 GPUs.

\subsubsection{Ablation Study}
To evaluate the contributions of each component in the proposed TiFRe framework and the impact of key hyperparameters, we conducted an ablation study on VNBench. As shown in Table~\ref{tab:ablation_without_k}, we examined the effects of different modules and configurations, including a detailed analysis of varying the maximum frame count parameter \( k \) (Table~\ref{tab:ablation_with_k}). Here, \( k \) represents the upper limit of selected frames, though the actual number may be less if fewer keyframes are detected.

1.\textbf{Baseline (No TiFRe)}: For comparison, we evaluate the base model without integrating TiFRe. The baseline uses fixed frame sampling at a pre-determined frame rate, which introduces redundancy and misses semantically critical frames.

2.\textbf{Without Text-Guided Frame Sampling (No TFS)}:  In this configuration, text-guided analysis is removed, and frame selection is based on a fixed frame rate strategy, with a reduction in the total number of frames. This approach disregards the semantic alignment with user prompts, significantly reducing the model's ability to select contextually relevant frames, especially for tasks that require precise reasoning based on textual cues.

3.\textbf{Without Frame Matching and Merging (No FMM)}: Here, the semantic merging step is excluded, and only the selected frames are used as input to the model. This omission results in information loss from unselected frames, reducing the overall accuracy.

4.\textbf{Full Model (TiFRe)}: This represents the complete implementation of TiFRe, incorporating both Text-guided Frame Sampling (TFS) and Frame Matching and Merging (FMM) to balance efficiency and semantic retention.

5.\textbf{Varying \textit{k} (Effect of Maximum Frame Count)}: We analyze the effect of changing the maximum frame count \textit{k} on performance. A larger \textit{k} allows more frames to be retained but increases computational cost, while a smaller \textit{k} reduces overhead but risks omitting critical information. This study evaluates the trade-offs between computational efficiency and performance across different values of \textit{k}.

\begin{figure*}[h]
    \centering
    \includegraphics[width=1\textwidth]{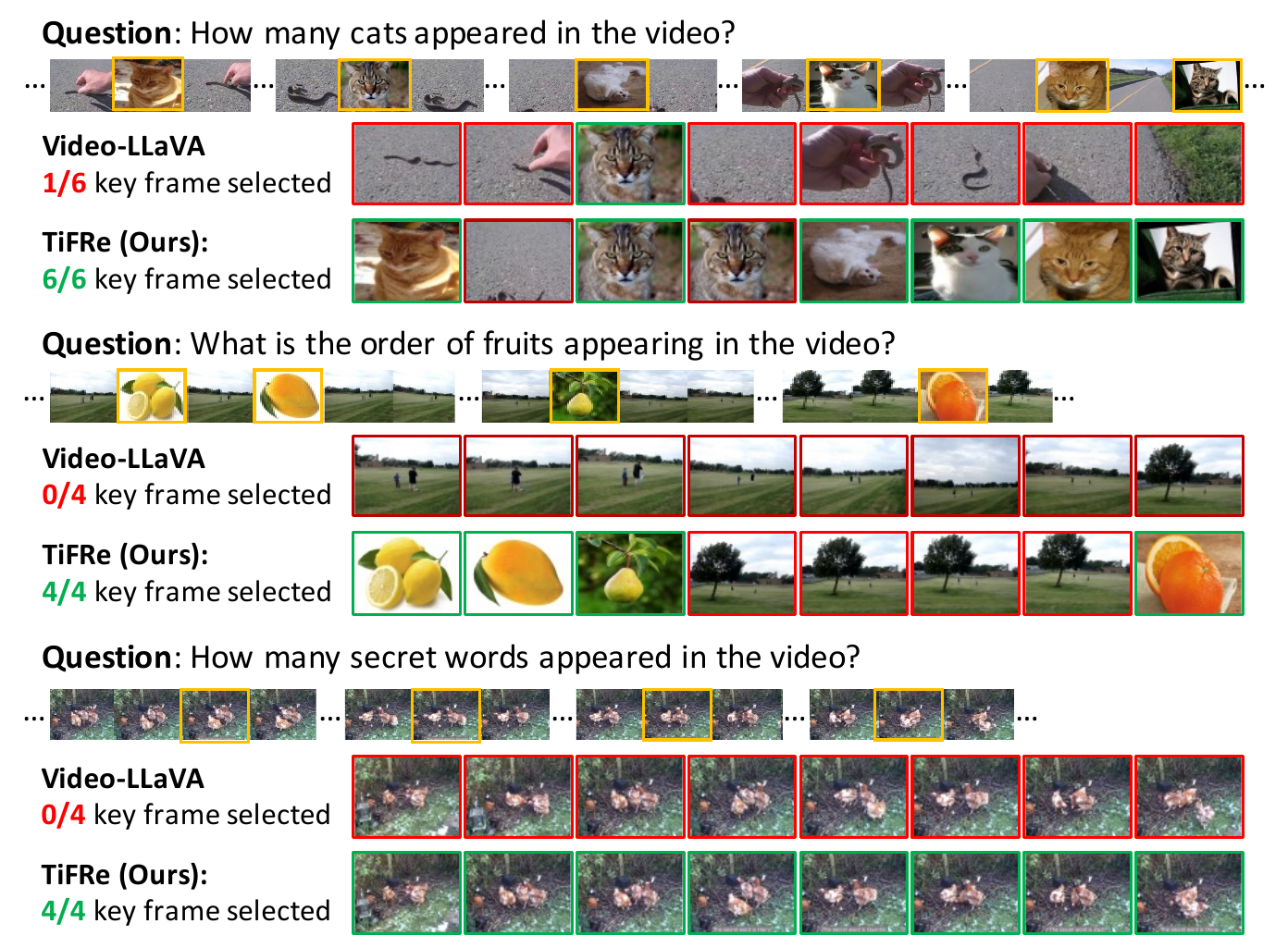}
    \caption{Examples of key frame selection. For each example, the first row is the raw video, where ground-truth key frames are highlighted with \textcolor{yellow}{yellow} boxes. The second row is key frames selected by Video-LLaVA and the third row is by TiFRe (ours), where key frames are highlighted with \textcolor{green}{green} boxes and other non-key frames are with \textcolor{red}{red} boxes.}
    \label{fig:example}
    \vspace{-0.3cm}
\end{figure*}

\subsection{Visualizations}


We present several key frame selection results in Figure \ref{fig:example}. In these examples, the input raw video sequence exceeds one minute in length, consisting of a large volume of frames to process. After applying a coarse-level key frame selection, the total number of frames is reduced to over 60. To highlight the frames that contain essential information related to the user’s question, we mark these critical frames with yellow boxes in the visualized sequence of input frames. In the examples provided, the first and third tasks involve counting, and the second example focuses on an ordering task. Notably, the third example presents an additional challenge, as it requires a fine-grained understanding of both the video content and the textual captions that accompany specific frames. For this example, frames containing captions are considered key frames due to their relevance to the question posed. Video-LLaVA, which selects frames at a fixed frame rate, fails to select most of the key frames in these sequences. In contrast, our TiFRe approach successfully identifies all the key frames from each raw video sequence.
\section{Conclusions}

In this paper, we proposed TiFRe, a text-guided frame reduction framework that significantly improves the efficiency and accuracy of Video MLLMs by dynamically selecting key frames based on user prompts and fusing information from non-selected frames. Through extensive experiments, TiFRe demonstrated its ability to reduce computational costs while achieving superior performance compared to state-of-the-art methods like Video-XL and Video-LLaVA. By effectively balancing computational efficiency and semantic preservation, TiFRe establishes itself as a robust solution for efficient video understanding tasks.

{
    \small
    \bibliographystyle{ieeenat_fullname}
    \bibliography{main}
}


\end{document}